\documentclass[11pt,a4paper]{article}
\usepackage[hyperref]{emnlp2018}
\usepackage{times}
\usepackage{latexsym}
\usepackage{booktabs}
\usepackage{graphicx}
\usepackage{graphbox}
\usepackage{amsmath}
\usepackage{amssymb}
\usepackage{url}

\usepackage{xargs} 
\usepackage[colorinlistoftodos,prependcaption,textsize=tiny]{todonotes}
\usepackage{marginnote}

\aclfinalcopy 

\title{Dynamic Meta-Embeddings for Improved Sentence Representations}

\author{Douwe Kiela$^{\dagger}$, Changhan Wang$^{\dagger}$ and Kyunghyun Cho$^{\dagger\ddagger\star}$ \\
  $^{\dagger}$ Facebook AI Research; $^\ddagger$ New York University; $^\star$ CIFAR Global Scholar
  \\
  {\tt \{dkiela,changhan,kyunghyuncho\}@fb.com}\\}

\date{}

\begin{document}
\maketitle
\begin{abstract}
While one of the first steps in many NLP systems is selecting what pre-trained word embeddings to use, we argue that such a step is better left for neural networks to figure out by themselves. To that end, we introduce dynamic meta-embeddings, a simple yet effective method for the supervised learning of embedding ensembles, which leads to state-of-the-art performance within the same model class on a variety of tasks. We subsequently show how the technique can be used to shed new light on the usage of word embeddings in NLP systems.
\end{abstract}

\section{Introduction}
It is no exaggeration to say that word embeddings have revolutionized NLP. From early distributional semantic models \cite{Turney:2010jair,Erk:2012llc,Clark:2015book} to deep learning-based word embeddings \cite{Bengio:2003jmlr,Collobert:2008icml,Mikolov:2013nips,Pennington:2014emnlp,Bojanowski:2016arxiv}, word-level meaning representations have found applications in a wide variety of core NLP tasks, to the extent that they are now ubiquitous in the field \cite{Goldberg:2016jair}.

A sprawling literature has emerged about what types of embeddings are most useful for which tasks. For instance, there has been extensive work on understanding what word embeddings learn \cite{Levy:2014nips}, evaluating their performance \cite{Milajevs:2014arxiv,Schnabel:2015emnlp,Bakarov:2017arxiv}, specializing them for certain tasks \cite{Maas:2011acl,Faruqui:2014arxiv,Kiela:2015emnlp,Mrksic:2016acl,Vulic:2017arxiv}, learning sub-word level representations \cite{Wieting:2016arxiv,Bojanowski:2016arxiv,Lee:2016arxiv}, et cetera.

One of the first steps in designing many NLP systems is selecting what kinds of word embeddings to use, with people often resorting to freely available pre-trained embeddings. While this is often a sensible thing to do, the usefulness of word embeddings for downstream tasks tends to be hard to predict, as downstream tasks can be poorly correlated with word-level benchmarks. An alternative is to try to combine the strengths of different word embeddings.  Recent work in so-called ``meta-embeddings'', which ensembles embedding sets, has been gaining traction \cite{Yin:2015arxiv,Bollegala:2017arxiv,Muromagi:2017nodalida,Coates:2018naacl}. Meta-embeddings are usually created in a separate preprocessing step, rather than in a process that is dynamically adapted to the task. In this work, we explore the supervised learning of task-specific, dynamic meta-embeddings, and
apply the technique to sentence representations.

The proposed approach turns out to be highly effective, leading to state-of-the-art performance within the same model class on a variety of tasks, opening up new areas for exploration and yielding insights into the usage of word embeddings.

\paragraph{Why Is This a Good Idea?} Our technique brings several important benefits to NLP applications. First, it is embedding-agnostic, meaning that one of the main (and perhaps most important) hyperparameters in NLP pipelines is made obsolete. Second, as we will show, it leads to improved performance on a variety of tasks. Third, and perhaps most importantly, it allows us to overcome common pitfalls with current systems:

\begin{itemize}
\item \textbf{Coverage} One of the main problems with NLP systems is dealing with out-of-vocabulary words: our method increases lexical coverage by allowing systems to take the union over different embeddings.
\item \textbf{Multi-domain} Standard word embeddings are often trained on a single domain, such as Wikipedia or newswire. With our method, embeddings from different domains can be combined, optionally while taking into account contextual information.
\item \textbf{Multi-modality} Multi-modal information has proven useful in many tasks \cite{Baroni:2016llc,Baltruvsaitis:2018pami}, yet the question of multi-modal fusion remains an open problem. Our method offers a straightforward solution for combining information from different modalities.
\item \textbf{Evaluation} While it is often unclear how to evaluate word embedding performance, our method allows for inspecting the weights that networks assign to different embeddings, providing a direct, task-specific, evaluation method for word embeddings.
\item \textbf{Interpretability and Linguistic Analysis} Different word embeddings work well on different tasks. This is well-known in the field, but knowing \emph{why} this happens is less well-understood. Our method sheds light on which embeddings are preferred in which linguistic contexts, for different tasks, and allows us to speculate as to why that is the case.
\end{itemize}

\paragraph{Outline} In what follows, we explore dynamic meta-embeddings and show that this method outperforms the naive concatenation of various word embeddings, while being more efficient. We apply the technique in a BiLSTM-max sentence encoder \cite{Conneau:2017emnlp} and evaluate it on well-known tasks in the field: natural language inference (SNLI and MultiNLI; \S\ref{sec:nli}), sentiment analysis (SST; \S\ref{sec:sst}), and image-caption retrieval (Flickr30k; \S\ref{sec:imgcap}). In each case we show state-of-the-art performance within the class of single sentence encoder models. Furthermore, we include an extensive analysis (\S\ref{sec:analysis}) to highlight the general usefulness of our technique and to illustrate how it can lead to new insights.

\section{Related Work}

Thanks to their widespread popularity in NLP, a sprawling literature has emerged about learning and applying word embeddings---much too large to fully cover here, so we focus on previous work that combines multiple embeddings for downstream tasks.

\newcite{Maas:2011acl} combine unsupervised embeddings with supervised ones for sentiment classification. \newcite{Yang:2017iclr} and \newcite{Miyamoto:2016arxiv} learn to combine word-level and character-level embeddings. Contextual representations have been used in neural machine translation as well, e.g. for learning contextual word vectors and applying them in other tasks \cite{Mccann:2017nips} or for learning context-dependent representations to solve disambiguation problems in machine translation \newcite{Choi:2016arxiv}.

Neural tensor skip-gram models learn to combine word, topic and context embeddings \cite{Liu:2015ijcai}; context2vec \cite{Melamud:2016nll} learns a more sophisticated context representation separately from target embeddings; and \newcite{Li:2016kbs} learn word representations with distributed word representation with multi-contextual mixed embedding. Recent work in ``meta-embeddings'', which ensembles embedding sets, has been gaining traction \cite{Yin:2015arxiv,Bollegala:2017arxiv,Muromagi:2017nodalida,Coates:2018naacl}---here, we show that the idea can be applied in context, and to sentence representations. Furthermore, these works obtain meta-embeddings as a preprocessing step, rather than learning them dynamically in a supervised setting, as we do here. Similarly to \newcite{Peters:2018arxiv}, who proposed deep contextualized word representations derived from language models and which led to impressive performance on a variety of tasks, our method allows for contextualization, in this case of embedding set weights.

There has also been work on learning multiple embeddings per word \cite{Chen:2014emnlp,Neelakantan:2015arxiv,Vu:2016naacl}, including a lot of work in sense embeddings where the senses of a word have their own individual embeddings \cite{Iacobacci:2015acl,Qiu:2016emnlp}, as well as on how to apply such sense embeddings in downstream NLP tasks \cite{Pilehvar:2017arxiv}.

The question of combining multiple word embeddings is related to multi-modal and multi-view learning. For instance, combining visual features from convolutional neural networks with word embeddings has been examined \cite{Kiela:2014emnlp,Lazaridou:2015arxiv}, see \newcite{Baltruvsaitis:2018pami} for an overview. In multi-modal semantics, for instance, word-level embeddings from different modalities are often mixed via concatenation $\mathbf{r} = [\alpha \mathbf{u}, (1-\alpha) \mathbf{v}]$ \cite{Bruni:2014jair}. Here, we dynamically learn the weights to combine representations. Recently, related dynamic multi-modal fusion methods have also been explored \cite{Wang:2018arxiv,Kiros:2018acl}. There has also been work on unifying multi-view embeddings from different data sources \cite{Luo:2014}.

The usefulness of different embeddings as initialization has been explored \cite{Kocmi:2017arxiv}, and different architectures and hyperparameters have been extensively examined \cite{Levy:2015tacl}. Problems with evaluating word embeddings intrinsically are well known \cite{Faruqui:2016arxiv}, and various alternatives for evaluating word embeddings in downstream tasks have been proposed \cite[e.g.,][]{Tsvetkov:2015emnlp,Schnabel:2015emnlp,Ettinger:2016repeval}. For more related work with regard to word embeddings and their evaluation, see \newcite{Bakarov:2017arxiv}.

Our work can be seen as an instance of the well-known attention mechanism \cite{Bahdanau:2014arxiv}, and its recent sentence-level incarnations of self-attention \cite{Lin:2017arxiv} and inner-attention \cite{Cheng:2016arxiv,Liu:2016arxiv}, where the attention mechanism is applied within the same sentence instead of for aligning multiple sentences. Here, we learn (optionally contextualized) attention weights for different embedding sets and apply the technique in sentence representations \cite{Kiros:2015nips,Wieting:2015arxiv,Hill:2016arxiv,Conneau:2017emnlp}.

\section{Dynamic Meta-Embeddings}

Commonly, NLP systems use a single type of word embedding, e.g., word2vec \cite{Mikolov:2013nips}, GloVe \cite{Pennington:2014emnlp} or FastText \cite{Bojanowski:2016arxiv}. We propose giving networks access to multiple types of embeddings, allowing a network to learn which embeddings it prefers by predicting a weight for each embedding type, optionally depending on the context.

For a sentence of $s$ tokens $\{\mathbf{t}_{j}\}_{j=1}^{s}$, we have $n$ word embedding types, leading to sequences $\{\mathbf{w}_{i, j}\}_{j=1}^{s}\in\mathbb{R}^{d_i}\:(i=1,2,\ldots,n)$. We center each type of word embedding to zero mean.

\paragraph{Naive baseline}

We compare to naive concatenation as a baseline. Concatenation is a sensible strategy for combining different embedding sets, because it provides the sentence encoder with all of the information in the individual embeddings:

\begin{equation*}
\mathbf{w}^{CAT}_{j} = [\mathbf{w}_{1,j}, \mathbf{w}_{2,j}, \ldots, \mathbf{w}_{n,j}].
\end{equation*}

\noindent The downside of concatenating embeddings and giving that as input to an RNN encoder, however, is that the network then quickly becomes inefficient as we combine more and more embeddings.

\paragraph{DME}

For dynamic meta-embeddings, we project the embeddings into a common $d'$-dimensional space by learned linear functions $\mathbf{w'}_{i,j}=\mathbf{P}_i\mathbf{w}_{i,j}+\mathbf{b}_i\:(i=1,2,\ldots,n)$ where $\mathbf{P}_i\in\mathbb{R}^{d'\times d_i}$ and $\mathbf{b}_i\in\mathbb{R}^{d'}$. We then combine the projected embeddings by taking the weighted sum 
\[
\mathbf{w}^{DME}_{j}=\sum_{i=1}^{n}\alpha_{i,j}\mathbf{w'}_{i,j}
\]
where $\alpha_{i,j}=g(\{\mathbf{w'}_{i,j}\}_{j=1}^{s})$ are scalar weights from a self-attention mechanism:
\begin{equation}
\label{eq:dme}
\alpha_{i,j}=g(\mathbf{w'}_{i,j})=\phi(\mathbf{a}\cdot\mathbf{w'}_{i,j} + b)
\end{equation}
where $\mathbf{a}\in\mathbb{R}^{d'}$ and $b\in\mathbb{R}$ are learned parameters and $\phi$ is a softmax (or could be a sigmoid or tanh, for gating). We also experiment with an \textbf{Unweighted} variant of this approach, that just sums up the projections.

\paragraph{CDME}

Alternatively, we can make the self-attention mechanism context-dependent, leading to contextualized DME (CDME):
\begin{equation}
\label{eq:cdme}
\alpha_{i,j}=g(\{\mathbf{w'}_{i,j}\}_{j=1}^{s})=\phi(\mathbf{a}\cdot\mathbf{h}_{j} + b)
\end{equation}
where $\mathbf{h}_{j}\in\mathbb{R}^{2m}$ is the $j^{th}$ hidden state of a BiLSTM taking $\{\mathbf{w'}_{i,j}\}_{j=1}^{s}$ as input, $\mathbf{a}\in\mathbb{R}^{2m}$ and $b\in\mathbb{R}$. We set $m=2$, which makes the contextualization very efficient.

\paragraph{Sentence encoder}

We use a standard bidirectional LSTM encoder with max-pooling (BiLSTM-Max), which computes two sets of $s$ hidden states, one for each direction:
\begin{align*}
\overrightarrow{\mathbf{h}_j} = \overrightarrow{\text{LSTM}_j}(\mathbf{w}_{1}, \mathbf{w}_{2}, \ldots, \mathbf{w}_{j})\\
\overleftarrow{\mathbf{h}_j} = \overleftarrow{\text{LSTM}}_j(\mathbf{w}_{j}, \mathbf{w}_{j+1}, \ldots, \mathbf{w}_{s})
\end{align*}

\noindent The hidden states are subsequently concatenated for each timestep to obtain the final hidden states, after which a max-pooling operation is applied over their components to get the final sentence representation:

\[
\mathbf{h} = \max(\{[\overrightarrow{\mathbf{h}_j}, \overleftarrow{\mathbf{h}_j}]\}_{j=1, 2, \ldots, s})
\]

\begin{table}[t]
  \centering
  \small
  \begin{tabular}{lrrr}
    \toprule
    Model & {\small SNLI} & {\small MNLI}\\
    \midrule
    {\small InferSent \cite{Conneau:2017emnlp}} & 84.5 & -\\
    {\small NSE \cite{Munkhdalai:2017acl}} & 84.6 & -\\
    {\small G-TreeLSTM \cite{Choi:2017arxiv}} & 86.0 & -\\
    {\small SSE \cite{Nie:2017arxiv}} & 86.1 & 73.6\\
    {\small ReSan \cite{Shen:2018arxiv}} & 86.3 & -\\
    \midrule
    GloVe BiLSTM-Max {\small (8.6M)} & 85.2$\pm$.3 & 70.0$\pm$.5\\
    FastText BiLSTM-Max {\small (8.6M)} & 85.2$\pm$.2 & 70.3$\pm$.3\\
    Naive baseline {\small (9.8M)} & 85.6$\pm$.3 & 71.1$\pm$.2\\
	Naive baseline {\small (61.3M)} & 86.0$\pm$.5 & 73.0$\pm$.2\\\midrule
    Unweighted DME {\small (8.6M)} & 86.3$\pm$.4 & 74.4$\pm$.2\\
    DME {\small (8.6M)} & 86.2$\pm$.2 & 74.4$\pm$.2\\
    CDME {\small (8.6M)} & 86.4$\pm$.3 & 74.1$\pm$.2\\\midrule
    DME* {\small (9.0M)} & 86.7$\pm$.2 & 74.3$\pm$.4\\
	CDME* {\small (9.0M)} & 86.5$\pm$.2 & 74.9$\pm$.5\\
    \bottomrule
  \end{tabular}
  \caption{\label{table:nli}Accuracy scores on the Stanford Natural Language Inference (SNLI) and MultiNLI Mismatched (MNLI) tasks. DME=Dynamic Meta-Embeddings; CDME=Contextualized Dynamic Meta-Embeddings;  *=multiple different embedding sets (see Section \ref{sec:nli}). Number of parameters included in parenthesis. Results averaged over five runs with different random seeds, using a BiLSTM-Max sentence encoder.}
\end{table}

\section{Natural Language Inference}
\label{sec:nli}

Natural language inference, also known as recognizing textual entailment (RTE), is the task of classifying pairs of sentences according to whether they are neutral, entailing or contradictive. Inference about entailment and contradiction is fundamental to understanding natural language, and there are two established datasets to evaluate semantic representations in that setting: SNLI \cite{Bowman:2015emnlp} and the more recent MultiNLI \cite{Williams:2017arxiv}.

The SNLI dataset consists of 570k human-generated English sentence pairs, manually labeled for entailment,
contradiction and neutral.  The MultiNLI dataset can be seen as an extension of SNLI: it contains 433k sentence pairs, taken from ten different genres (e.g. fiction, government text or spoken telephone conversations), with the same entailment labeling scheme.

We train sentence encoders with dynamic meta-embeddings using two well-known and often-used embedding types: FastText \cite{Mikolov:2018lrec,Bojanowski:2016arxiv} and GloVe \cite{Pennington:2014emnlp}. Specifically, we make use of the 300-dimensional embeddings trained on a similar WebCrawl corpus, and compare three scenarios: when used individually, when naively concatenated or in the dynamic meta-embedding setting (unweighted, context-independent DME and contextualized CDME). We also compare our approach against other models in the same class---in this case, models that encode sentences individually and do not allow attention \emph{across} the two sentences.\footnote{This is a common distinction, see e.g. the SNLI leaderboard at \url{https://nlp.stanford.edu/projects/snli/}.} We include InferSent \cite{Conneau:2017emnlp}, which also makes use of a BiLSTM-Max sentence encoder.

In addition, we include a setting where we combine not two, but six different embedding types, adding FastText wiki-news embeddings\footnote{See \url{https://fasttext.cc/}}, English-German and English-French embeddings from \newcite{Hill:2014nmt}, as well as the BOW2 embeddings from \newcite{Levy:2014acl} trained on Wikipedia.

\subsection{Implementation Details}

The two sentences are represented individually using the sentence encoder. As is standard in the literature, the sentence representations are subsequently combined using $\mathbf{m} = [\mathbf{u}, \mathbf{v}, \mathbf{u}*\mathbf{v},|\mathbf{u}-\mathbf{v}|]$. We train a two-layer classifier with rectifiers on top of the combined representation. Notice that there is no interaction (e.g., attention) between the representations of $\mathbf{u}$ and $\mathbf{v}$ for this class of model.

We use 256-dimensional embedding projections,  512-dimensional BiLSTM encoders and an MLP with 1024-dimensional hidden layer in the classifier. The initial learning rate is set to $0.0004$ and dropped by a factor of $0.2$ when dev accuracy stops improving, dropout to $0.2$, and we use Adam for optimization \cite{Kingma:2014arxiv}. The loss is standard cross-entropy.

For MultiNLI, which has no designated validation set, we use the in-domain \emph{matched} set for validation and report results on the out-of-domain \emph{mismatched} set.

\subsection{Results}

Table \ref{table:nli} shows the results. We report accuracy scores averaged over five runs with different random seeds, together with their standard deviation, for the SNLI and MultiNLI datasets. We include two versions of the naive baseline: one with a 512-dimensional BiLSTM encoder; and a bigger one with 2048 dimensions. Both naive baseline models outperform the single encoders that have only GloVe or FastText embeddings. This shows how including more than one embeddings can help performance. Next, we observe that the DME embeddings outperform the naive concatenation baselines, while having fewer parameters. Differences between the three DME variants are small and not significant, although we do note that we found the highest maximum performance for the contextualized version, which adds very few additional parameters. It is important to note that the imposition of weighting thus is not detrimental to performance, which means that DME and CDME provide additional interpretability without sacrificing performance.

Finally, we obtain results for using the six different embedding types (marked *), and show that adding in more embeddings increases performance further. To our knowledge, these numbers constitute the state of the art within the model class of single sentence encoders on these tasks.

\begin{table}[t]
  \centering
  \begin{tabular}{lr}
    \toprule
    Model & SST\\
    \midrule
    Const. Tree LSTM \cite{Tai:2015arxiv} & 88.0\\
    DMN \cite{Kumar:2016icml} & 88.6\\
    DCG \cite{Looks:2017arxiv} & 89.4\\
    NSE \cite{Munkhdalai:2017acl} & 89.7\\
    \midrule
    GloVe BiLSTM-Max {\small (4.1M)} & 88.0$\pm$.1\\
	FastText BiLSTM-Max {\small (4.1M)} & 86.7$\pm$.3\\
    Naive baseline {\small (5.4M)} & 88.5$\pm$.4\\\midrule
    Unweighted DME {\small (4.1M)} & 89.0$\pm$.2\\
    DME {\small (4.1M)} & 88.7$\pm$.6\\
    CDME {\small (4.1M)} & 89.2$\pm$.4\\\midrule
    CDME*-Softmax {\small (4.6M)} & 89.3$\pm$.5\\
	CDME*-Sigmoid {\small (4.6M)} & 89.8$\pm$.4\\
    \bottomrule
  \end{tabular}
  \caption{\label{table:sentiment}Sentiment classification accuracy results on the binary SST task. For DCG we compare against their best single sentence model \cite{Looks:2017arxiv}. *=multiple different embedding sets (see Section \ref{sec:nli}). Number of parameters included in parenthesis. Results averaged over ten runs with different random seeds.}
\end{table}
  
\section{Sentiment}
\label{sec:sst}

To showcase the general applicability of the proposed approach, we also apply it to a case where we have to classify a single sentence, namely, sentiment classification. Sentiment analysis and opinion mining have become important applications for NLP research. We evaluate on the binary SST task \cite{Socher:2013emnlp}, consisting of 70k sentences with a corresponding binary (positive or negative) sentiment label.

\subsection{Implementation Details}

We use 256-dimensional embedding projections, 512-dimensional BiLSTM encoders and an MLP with 512-dimensional hidden layer in the classifier. The initial learning rate is set to $0.0004$ and dropped by a factor of $0.2$ when dev accuracy stops improving, dropout to $0.5$, and we use Adam for optimization. The loss is standard cross-entropy. We calculate the mean accuracy and standard deviation based on ten random seeds.

\subsection{Results}
Table \ref{table:sentiment} shows a similar pattern as we observed with NLI: the naive baseline outperforms the single-embedding encoders; the DME methods outperform the naive baseline, with the contextualized version appearing to work best. Finally, we experiment with replacing $\phi$ in Eq. \ref{eq:dme} and \ref{eq:cdme} with a sigmoid gate instead of a softmax, and observe improved performance on this task, outperforming the comparable models listed in the table. These results further strengthen the point that having multiple different embeddings helps, and that we can learn to combine those different embeddings efficiently, in interpretable ways.

\begin{table}[t]
  \centering
  \small
  \begin{tabular}{lcccc}
    \toprule
    & \multicolumn{2}{c}{Image} & \multicolumn{2}{c}{Caption}\\
    Model $|  $ R@: & 1 & 10 & 1 & 10\\
    \midrule
    VSE++ & 32.3 & 72.1 & 43.7 & 82.1\\\midrule
	FastText (15M) & 35.6 & 74.7 & 47.1 & 82.7 \\
	ImageNet (29M) & 25.6 & 63.1 & 36.6  & 72.2 \\
	Naive (32M) & 34.4 & 73.9 & 46.4 & 82.2\\\midrule
	Unweighted DME (15M) & 35.9 & 75.0 & 48.9 & 83.7\\
    DME (15M) & 36.5 & 75.5 & 49.7 & 83.6\\
    CDME (15M) & 36.5 & 75.6 & 49.0 & 83.8\\
    \bottomrule
  \end{tabular}
  \caption{\label{table:imgcap}Image and caption retrieval results (R@1 and R@10) on Flickr30k dataset, compared to VSE++ baseline \cite{Faghri:2017arxiv}. VSE++ numbers in the table are with ResNet features and random cropping, but no fine-tuning. Number of parameters included in parenthesis; averaged over five runs with std omitted for brevity.}
\end{table}

\section{Image-Caption Retrieval}
\label{sec:imgcap}

An advantage of the proposed approach is that it is inherently capable of dealing with multi-modal information. Multi-modal semantics \cite{Bruni:2014jair} often combines linguistic and visual representations via concatenation with a global weight $\alpha$, i.e., $\mathbf{v} = [\alpha \mathbf{v}_{ling}, (1-\alpha) \mathbf{v}_{vis}]$. In DME we instead learn to combine embeddings dynamically, optionally based on context. The representation for a word then becomes grounded in the visual modality, and we encode on the word-level \emph{what things look like}.

We evaluate this idea on the Flickr30k image-caption retrieval task: given an image, retrieve the correct caption; and vice versa. The intuition is that knowing what something looks like makes it easier to retrieve the correct image/caption. While this work was under review, a related method was published by \newcite{Kiros:2018acl}, which takes a similar approach but evaluates its effectiveness on COCO and uses Google images. We obtain word-level \emph{visual embeddings} by retrieving relevant images for a given label from ImageNet in the same manner as \newcite{Kiela:2014emnlp}, taking the images' ResNet-152 features \cite{He:2016cvpr} and subsequently averaging those. We then learn to combine textual (FastText) and visual (ImageNet) word representations in the caption encoder used for retrieving relevant images.

\subsection{Implementation Details}

Our loss is a max-margin rank loss as in VSE++ \cite{Faghri:2017arxiv}, a state-of-the-art method on this task. The network architecture is almost identical to that system, except that we use DME (with 256-dimensional embedding projection) and a 1024-dimensional caption encoder. For the Flickr30k images that we do retrieval over, we use random cropping during training for data augmentation and use a ResNet-152 for feature extraction. We tune the sizes of the encoders and use a learning rate of $0.0003$ and a dropout rate of $0.1$.

\subsection{Results}

Table \ref{table:imgcap} shows the results, comparing against VSE++. First, note that the ImageNet-only embeddings don't work as well as the FastText ones, which is most likely due to poorer coverage. We observe that DME outperforms naive and FastText-only, and outperforms VSE++ by a large margin. These findings confirm the intuition that knowing what things look like (i.e., having a word-level visual representation) improves performance in visual retrieval tasks (i.e., where we need to find relevant images for phrases or sentences)---something that sounds obvious but has not really been explored before, to our knowledge. This showcases DME's usefulness for fusing embeddings in multi-modal tasks.

\section{Discussion \& Analysis}
\label{sec:analysis}

Aside from improved performance, an additional benefit of learning dynamic meta-embeddings is that they enable inspection of the weights that the network has learned to assign to the respective embeddings. In this section, we perform a variety of smaller experiments in order to highlight the usefulness of the technique for studying linguistic phenomena, determining appropriate training domains and evaluating word embeddings. We compute the contribution of each word embedding type as follows:

\[
\beta_{i,j} = \frac{\Vert \alpha_{i,j} \mathbf{w'}_{i,j} \Vert_2}{\sum_{k=1}^n \Vert \alpha_{k,j} \mathbf{w'}_{k,j} \Vert_2}
\]

\begin{figure*}[t]
\centering
\includegraphics[width=\textwidth]{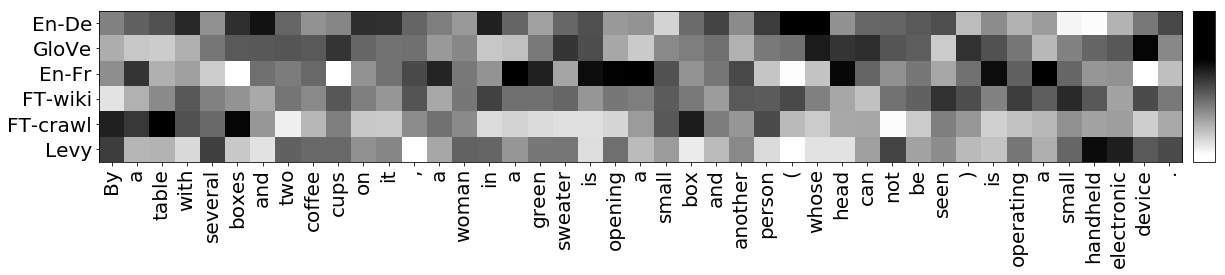}
\caption{Example visualization of a sentence from the SNLI dev set.}
\label{fig:visualization-example}
\end{figure*}

\begin{figure*}[t]
\centering
\includegraphics[width=\columnwidth]{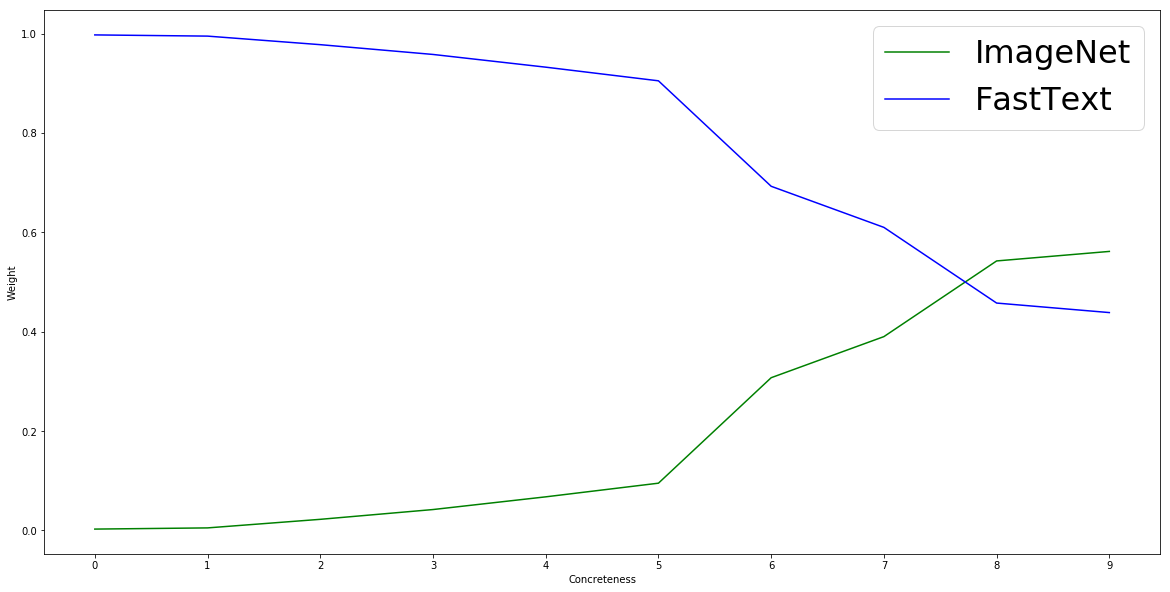}
\includegraphics[width=\columnwidth]{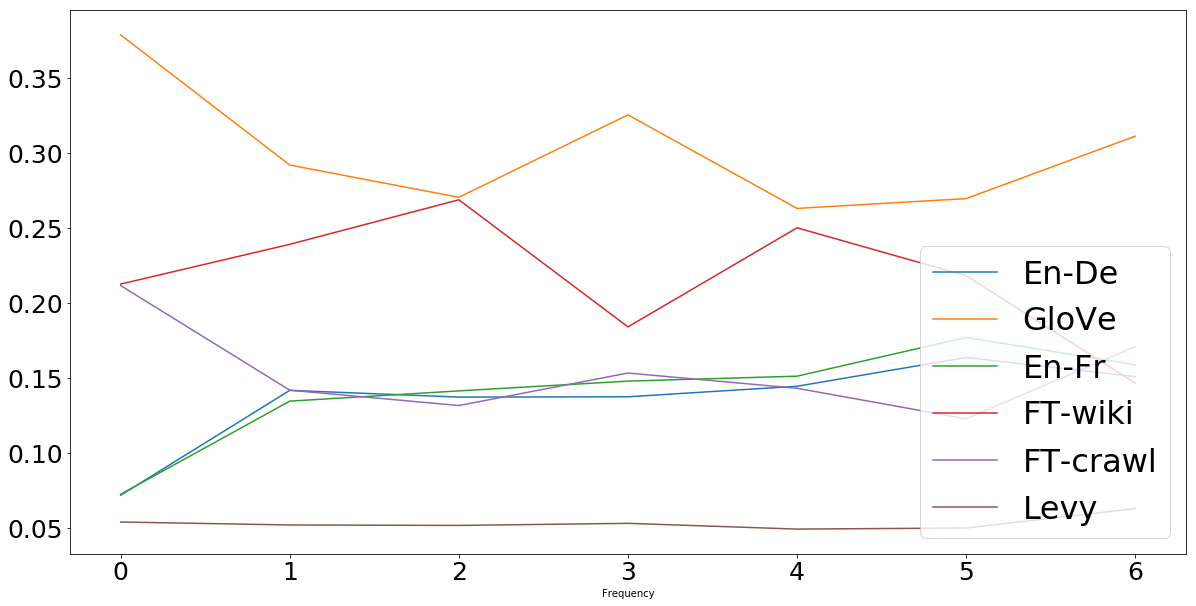}
\caption{Concreteness weights (left) for Flickr30k model and Frequency weights (right) for SNLI model with multiple embeddings. Visual ImageNet embeddings are preferred for more concrete words. GloVe is strongly preferred for low-frequency words.}
\label{fig:freq-and-conc}
\end{figure*}

\begin{figure}[t]
\centering
\includegraphics[width=\columnwidth]{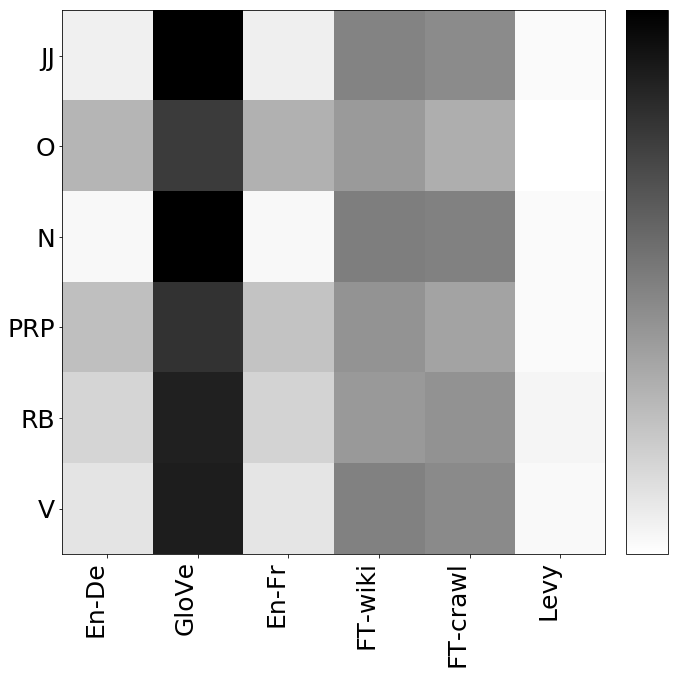}
\caption{Pos tags and weights by embedding type.}
\label{fig:postags}
\end{figure}

\begin{figure}[t]
\centering
\includegraphics[width=\columnwidth]{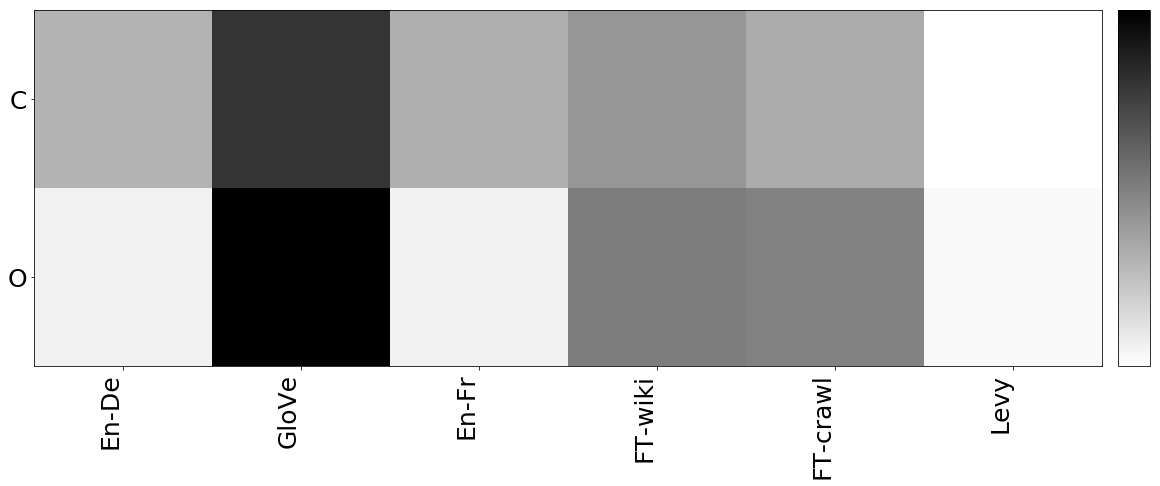}
\caption{Weights for open/closed class words.}
\label{fig:openclose}
\end{figure}

\begin{figure}[t]
\centering
\includegraphics[width=\columnwidth]{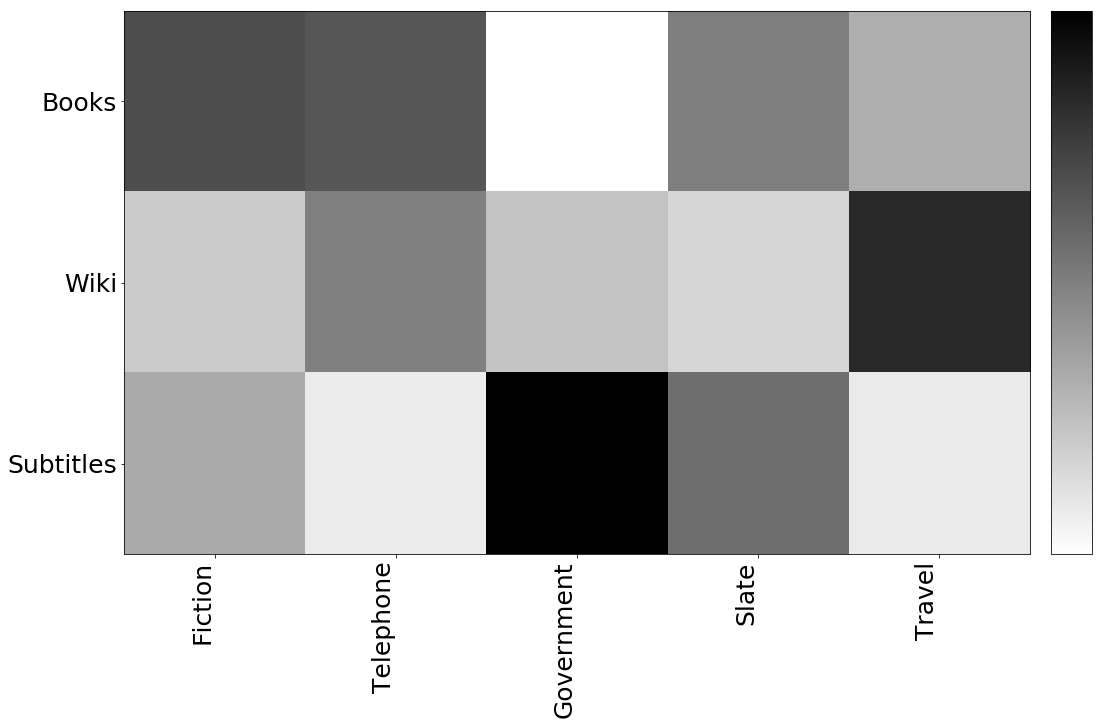}
\caption{Multi-domain weights on MultiNLI.}
\label{fig:multidomain}
\end{figure}

\subsection{Visualizing Attention}

Figure \ref{fig:visualization-example} shows the attention weights for a CDME model trained on SNLI, using the aforementioned six embedding sets. The sentence is from the SNLI validation set. We observe that different embeddings are preferred for different words. The figure is meant to illustrate possibilities for analysis, which we turn to in the next section.

\subsection{Linguistic Analysis}

We perform a fine-grained analysis of the behavior of DME on the validation set of SNLI. Figure \ref{fig:postags} shows a breakdown of the average attention weights per part of speech. Figure \ref{fig:openclose} shows a similar breakdown for open versus closed class. The analysis allows us to make several interesting observations: it appears that this model prefers GloVe embeddings, followed by the two FastText embeddings (trained on Wikipedia and Common Crawl). For open class words (e.g., nouns, verbs, adjectives and adverbs), those three embedding types are strongly preferred, while closed class words get more evenly divided attention. The embeddings from \newcite{Levy:2014acl} get low weights, possibly because the method is complementary with FastText-wiki, which was trained on a more recent version of Wikipedia.

We can further examine the attention weights by analyzing them in terms of frequency and concreteness. We use Norvig's Google N-grams corpus frequency counts\footnote{\url{http://norvig.com/mayzner.html}} to divide the words into frequency bins. Figure \ref{fig:freq-and-conc} (right) shows the average attention weights per frequency bin, ranging from low to high. We observe a clear preference for GloVe, in particular for low-frequency words.  For concreteness, we use the concreteness ratings from \newcite{Brysbaert:2014brm}. Figure \ref{fig:freq-and-conc} (left) shows the average weights per concreteness bin for a model trained on Flickr30k. We can clearly see that visual embeddings get higher weights as the words become more and more concrete.

There are of course intricate relationships between concreteness, frequency, POS tags and open/closed class words: closed class words are often frequent and abstract, while open class words could be more concrete, etc. It is beyond the scope of the current work to explore these further, but we hope that others will pursue this direction in future work.

\subsection{Multi-domain Embeddings}

The MultiNLI dataset consists of various genres. This allows us to inspect the applicability of source domain data for a specific genre. We train embeddings on three kinds of data: Wikipedia, the Toronto Books Corpus \cite{Zhu:2015iccv} and the English OpenSubtitles\footnote{\url{http://opus.nlpl.eu/OpenSubtitles.php}}. We examine the attention weights on the five genres in the in-domain (matched) set, consisting of fiction; transcriptions of spoken telephone conversations; government reports, speeches, letters and press releases; popular culture articles from the Slate Magazine archive; and travel guides.

Figure \ref{fig:multidomain} shows the average attention weights for the three embedding types over the five genres. We observe that Toronto Books, which consists of fiction, is very appropriate for the fiction genre, while Wikipedia is highly preferred for the travel genre, perhaps because it contains a lot of factual information about geographical locations. The government genre makes more use of OpenSubtitles. The spoken telephone genre does not appear to prefer OpenSubtitles, which we might have expected given that that corpus would contain spoken dialogue, but Toronto books, which does include written dialogue.

\begin{table}[t]
  \centering
  \begin{tabular}{lrr|r}
    \toprule
    Model & Levy & LEAR & SNLI\\
    \midrule
    CDME & 0.33 & 0.67 & 85.3$\pm$.9\\
	& & & \\
    \toprule
    Model & GloVe & Refined & SST\\
    \midrule
    CDME & 0.59 & 0.41 & 89.0$\pm$.4\\
    \bottomrule
  \end{tabular}
  \caption{\label{table:specialization} Accuracy and learned weights on SNLI using LEAR \cite{Vulic:2017arxiv} or SST using sentiment-refined embeddings using the specialization method from \newcite{Yu:2017emnlp}.}
\end{table}

\subsection{Specialization}

The above shows that we can use DME to analyze different embeddings on a task. Given the recent interest in the community in specializing, retro-fitting and counter-fitting word embeddings for given tasks, we examine whether the lexical-level benefits of specialization extend to sentence-level downstream tasks. After all, one of the main motivations behind work on lexical entailment is that it allows for better downstream textual entailment. Hence, we take the LEAR embeddings by \newcite{Vulic:2017arxiv}, which do very well on the HyperLex lexical entailment evaluation dataset \cite{Vulic:2017cl}. We compare their best-performing embeddings against the original embeddings that were used for specialization, derived from the BOW2 embeddings of \newcite{Levy:2014acl}. Similarly, we use the technique of \newcite{Yu:2017emnlp} for refining GloVe embeddings for sentiment, and evaluate model performance on the SST task.

Table \ref{table:specialization} shows that LEAR embeddings get high weights compared to the original source embeddings (``Levy'' in the table). Our analysis showed that LEAR was particularly favored for verbs (with average weights of $0.75$). The sentiment-refined embeddings were less useful, with the original GloVe embeddings receiving higher weights. These preliminary experiments show how DME models can be used for analyzing the performance of specialized embeddings in downstream tasks.

Note that different weighting mechanisms might give different results---we found that the normalization strategy and the depth of the network significantly influenced weight assignments in our experiments with specialized embeddings.

\subsection{Examining Contextualization}

We examined models trained on SNLI and looked at the variance of the attention weights per word in the dev set. If contextualization is important for getting the classification decision correct, then we would expect big differences in the attention weights per word depending on the context. Upon examination, we only found relatively few differences. In part, this may be explained by the small size of the dev set, but for the Glove+FastText model we inspected there were only around twenty words with any variance at all, which suggests that the field needs to work on more difficult semantic benchmark tasks. The words, however, where characterized by their polysemy, in particular by having both noun and verb senses. The following words were all in the top 20 most context-dependent words: \emph{mob}, \emph{boards}, \emph{winds}, \emph{trains}, \emph{pitches}, \emph{camp}.

\section{Conclusion}

We argue that the decision of which word embeddings to use in what setting should be left to the neural network. While people usually pick one type of word embeddings for their NLP systems and then stick with it, we find that dynamically learned meta-embeddings lead to improved results. In addition, we showed that the proposed mechanism leads to better interpretability and insightful linguistic analysis. We showed that the network learns to select different embeddings for different data, different domains and different tasks. We also investigated embedding specialization and examined more closely whether contextualization helps. To our knowledge, this work constitutes the first effort to incorporate multi-modal information on the language side of image-caption retrieval models; and the first attempt at incorporating meta-embeddings into large-scale sentence-level NLP tasks.

In future work, it would be interesting to apply this idea to different tasks, in order to explore what kinds of embeddings are most useful for core NLP tasks, such as tagging, chunking, named entity recognition, parsing and generation. It would also be interesting to further examine specialization and how it transfers to downstream tasks. Using this method for evaluating word embeddings in general, and how they relate to sentence representations in particular, seems a fruitful direction for further exploration. In addition, it would be interesting to explore how the attention weights change during training, and if, e.g., introducing entropy regularization (or even negative entropy) might improve results or interpretability further.

\section*{Acknowledgments}

We thank the anonymous reviewers for their comments. We also thank Marcus Rohrbach, Laurens van der Maaten, Ivan Vuli\'c, Edouard Grave, Tomas Mikolov and Maximilian Nickel for helpful suggestions and discussions with regard to this work.

\bibliography{acl2018}
\bibliographystyle{acl_natbib_nourl}

\end{document}